\documentclass{article}
\usepackage{hyperref}
\usepackage{graphicx}
\usepackage{amsmath, amssymb, amsthm, amscd}
\newcommand{\ra}{\rightarrow}

\newcommand{\cx}{\sqsubseteq}
\newcommand{\Con}{\textsc{con}}
\newcommand{\con}{\textsf{con}}
\newcommand{\glb}{\textsf{glb}}
\newcommand{\lub}{\textsf{lub}}

\newcommand{\dfn}{\textsf{def}}

\begin{document}

\newtheorem{lem}{Lemma}[section]
\newtheorem{pro}{Proposition}[section]
\newtheorem{thm}{Theorem}[section]
\newtheorem{cor}{Corollary}[section]

\theoremstyle{definition}
\newtheorem{ex}{Example}[section]
\newtheorem{defi}{Definition}[section]
\newtheorem{scsp}{Soft Problem}
\theoremstyle{remark}
\newtheorem{rmk}{Remark}[section]
\newtheorem{notation}{Notation}[section]
%\tableofcontents

\title{Soft constraint abstraction based on
 semiring homomorphism
\thanks{Work partially supported by
National Nature Science Foundation of China (60673105,60621062,
60496321).}}
\author{Sanjiang Li and Mingsheng Ying
\thanks{lisanjiang{@}tsinghua.edu.cn (S. Li),
yingmsh{@}tsinghua.edu.cn (M. Ying)}\\
  Department of Computer Science and Technology\\
Tsinghua University, Beijing 100084, China}

\date{}
\maketitle

\begin{abstract}
The semiring-based constraint satisfaction problems (semiring
CSPs), proposed by Bistarelli, Montanari and Rossi \cite{BMR97},
is a very general framework of soft constraints. In this paper we
propose an abstraction scheme for soft constraints that uses
semiring homomorphism. To find optimal solutions of the concrete
problem, the idea is, first working in the abstract problem and
finding its optimal solutions, then using them to solve the
concrete problem.

In particular, we show that a mapping preserves optimal solutions
if and only if it is an order-reflecting semiring homomorphism.
Moreover, for a semiring homomorphism $\alpha$ and a problem $P$
over $S$, if $t$ is optimal in $\alpha(P)$, then there is an
optimal solution $\bar{t}$ of $P$ such that $\bar{t}$ has the same
value as $t$ in $\alpha(P)$.

\vskip 2mm

\noindent{\it Keywords}: Abstraction; Constraint solving; Soft
constraint satisfaction; Semiring homomorphism; Order-reflecting.
\end{abstract}

\section{Introduction}
In the recent years there has been a growing interest in soft
constraint satisfaction. Various extensions of the classical
constraint satisfaction problems (CSPs) \cite{Mon74,Mac92} have
been introduced in the literature, e.g., Fuzzy CSP
\cite{Ros76,Dub93,Rut94}, Probabilistic CSP \cite{Far93}, Weighted
CSP \cite{Shapiro81,Fre92}, Possibilistic CSP \cite{Schiex92}, and
Valued CSP \cite{Schiex95}. Roughly speaking, these extensions are
just like classical CSPs except that each assignment of values to
variables in the constraints is associated to an element taken
from a semiring. Furthermore, nearly all of these extensions, as
well as classical CSPs, can be cast by the semiring-based
constraint solving framework, called SCSP (for \emph{Semiring
CSP}), proposed by Bistarelli, Montanari and Rossi \cite{BMR97}.

Compared with classical CSPs, SCSPs are usually more difficult to
process and to solve. This is mainly resulted by the complexity of
the underlying semiring structure. Thus working on a simplified
version of the given problem would be worthwhile. Given a concrete
SCSP, the idea is to get an abstract one by changing the semiring
values of the constraints without changing the structure of the
problem. Once the abstracted version of a given problem is
available, one can first process the abstracted version and then
bring back the information obtained to the original problem. The
main objective is to find an optimal solution, or a reasonable
estimation of it, for the original problem.

The translation from a concrete problem to its abstracted version is
established via a mapping between the two semirings. More
concretely, suppose $P$ is an SCSP over $S$, and $\widetilde{S}$ is
another semiring (possibly simpler than $S$). Given a mapping
$\alpha:S\ra\widetilde{S}$, we can translate the concrete problem
$P$ to another problem, $\alpha(P)$, over $\widetilde{S}$ in a
natural way. We then ask when is an optimal solution of the concrete
problem $P$ also optimal in the abstract problem $\alpha(P)$? and,
given an optimal solution of $\alpha(P)$, when and how can we find a
reasonable estimation for an optimal solution of $P$?

The answers to these questions will be helpful in deriving useful
information on the abstract problem and then taking some useful
information back to the concrete problem. This paper is devoted to
the investigation of the above questions.

These questions were first studied in Bistarelli, Codognet and Rossi
\cite{BCR02}, where they established a Galois insertion-based
abstraction framework for soft constraint problems. In particular,
they showed that \cite[Theorem~27]{BCR02} if $\alpha$ is an
\emph{order-preserving} Galois insertion, then optimal solutions of
the concrete problem are also optimal in the abstract problem. This
sufficient condition, however, turns out to be equivalent to say
$\alpha$ is a semiring isomorphism (see Proposition~\ref{prop:op}),
hence too restrictive. Theorem~29 of \cite{BCR02} concerns computing
bounds that approximate an optimal solution of the concrete problem.
The statement of this theorem as given there is incorrect since a
counter-example (see Soft Problem~\ref{scsp:thm29} in this paper)
shows that the result holds conditionally.

This paper shows that semiring homomorphism plays an important
role in soft constraint abstraction. More precisely, we show that
(Theorem~\ref{mythm27}) a mapping preserves optimal solutions if
and only if it is an order-reflecting semiring homomorphism, where
a mapping $\alpha:S\ra\widetilde{S}$ is \emph{order-reflecting} if
for any two $a,b\in S$, we have $a<_S b$ from
$\alpha(a)<_{\widetilde{S}}\alpha(b)$. Moreover, for a semiring
homomorphism $\alpha$ and a problem $P$ over $S$, if $t$ is
optimal in $\alpha(P)$, then there is an optimal solution
$\bar{t}$ of $P$ such that $\bar{t}$ has the same value as $t$ in
$\alpha(P)$ (see Theorem~\ref{mythm31}).

This paper is organized as follows. First, in Section~2 we give a
summary of the theory of soft constraints. The notion of
$\alpha$-\emph{translation} of semiring CSPs is introduced in
Section~3, where we show that $\alpha$ preserves problem ordering if
and only if $\alpha$ is a semiring homomorphism. Section~4 discusses
when a translation $\alpha$ preserves optimal solutions, i.e. when
all optimal solutions of the concrete problem are also optimal in
the abstract problem. In Section~5, we discuss, given an optimal
solution of the abstract problem, what we can say about optimal
solutions of the concrete problem. Conclusions are given in the
final section.

\section{Semiring Constraint Satisfaction Problem}
In this section we introduce several basic notions used in this
paper. In particular, we give a brief summary of the theory of
c-semiring based constraint satisfaction problem raised in
\cite{BMR97} (Bistarelli, Montanari and Rossi 1997). The notion of
semiring homomorphism is also introduced.

\subsection{c-semirings}

\begin{defi}[semirings and c-semirings \cite{BCR02}]
 A semiring is a tuple $S=\langle S, +, \times, {\bf 0}, {\bf 1}\rangle$
such that:
\begin{itemize}
\item [1.] $S$ is a set and ${\bf 0},{\bf 1}\in S$;

\item [2.] $+$ is commutative, associative and {\bf 0} is its unit
element;

\item [3.] $\times$ is associative, distributive over $+$, {\bf 1}
is its unit element and {\bf 0} is its absorbing element.
\end{itemize}
We call $+$ and $\times$, respectively, the \emph{sum} and the
\emph{product} operation. A c-semiring is a semiring $\langle S,
+, \times, {\bf 0}, {\bf 1}\rangle$ such that:
\begin{itemize}
\item [4.] $+$ is idempotent, {\bf 1} is its absorbing element,
and $\times$ is commutative.
\end{itemize}
\end{defi}
Consider the relation $\leq_S$ defined over $S$ such that $a\leq_S
b$ iff $a+b=b$. Then it is possible to prove that \cite{BMR97}:
\begin{itemize}
\item $\langle S,\leq_S\rangle$ is a lattice, {\bf 0} is its
bottom and {\bf 1} its top;

\item $+$ is the \lub\ (lowest upper bound) operator $\vee$ in the
lattice $\langle S,\leq_S\rangle$;

\item $\times$ is monotonic on $\leq_S$;

\item If $\times$ is idempotent, that is $a\times a=a$ for each
$a\in S$, then $\langle S,\leq_S\rangle$ is a distributive lattice
and $\times$ is its \glb\ (greatest lower bound) $\wedge$.
\end{itemize}
\begin{rmk}The above definition of c-semiring differs from
the one given in \cite{BMR97} simply in that a c-semiring, with
the induced partial order, is not necessarily complete. For
example, suppose ${\mathbb Q}$ is the set of rational number and
$S=[0,1]\cap{\mathbb Q}$ is the subalgebra of the fuzzy semiring
$S_{FCSP}=\langle[0,1],\vee,\wedge,0,1\rangle$. Then $S$ is a
c-semiring but $\langle S,\leq_S\rangle$ is not a complete
lattice, where $\leq_S$ is the partial order induced by the
semiring $S$, which happens to be the usual total order on $S$.
\end{rmk}

\subsection{Semiring homomorphism}

\begin{defi}[homomorphism]
 A mapping $\psi$ from semiring $\langle S, +, \times$,
{\bf 0}, {\bf 1}$\rangle$ to semiring $\langle \widetilde {S},
\widetilde {+}, \widetilde{\times}, \widetilde{\bf 0},
\widetilde{\bf 1}\rangle$ is said to be a semiring
\emph{homomorphism} if for any $a,b\in S$
\begin{itemize}
\item $\psi(\bf 0)=\widetilde{\bf 0}$, $\psi(\bf 1)=\widetilde{\bf
1}$; and

\item $\psi(a+b)=\psi(a)\widetilde {+}\psi(b)$; and

\item $\psi(a\times b)=\psi(a)\widetilde {\times}\psi(b)$.
\end{itemize}
A semiring homomorphism $\psi$ is said to be a semiring
\emph{isomorphism} if $\psi$ is a bijection. Note that a semiring
isomorphism is also an order isomorphism w.r.t. the induced
partial orders.
\end{defi}
We give some examples of semiring homomorphism.

\begin{ex}\label{ex:tom}
Let $S$ and $\widetilde{S}$ be two c-semirings such that
\begin{itemize}
\item [(i)] both $\leq_S$ and $\leq_{\widetilde{S}}$ are totally
ordered; and

\item[(ii)] both $\times$ and $\widetilde{\times}$ are idempotent,
i.e. both are \glb\ operators.
\end{itemize}
Then a monotonic mapping $\alpha:S\ra\widetilde{S}$ is a
homomorphism if and only if $\alpha(\bf 0)=\widetilde{\bf 0}$, and
$\alpha(\bf 1)=\widetilde{\bf 1}$.
\end{ex}

Recall that a congruence relation $\thicksim$ over a semiring $S$
is an equivalence relation that satisfies:
\begin{itemize}

\item[]if $a\thicksim a^\prime$ and $b\thicksim b^\prime$, then
$a+b\thicksim a^\prime+b^\prime$, and $a\times b\thicksim
a^\prime\times b^\prime$.
\end{itemize}
We write $S/\thicksim$ for the resulted quotient structure.

\begin{ex}[natural homomorphism]\label{ex:nat}
Suppose $S$ is a (c-)semiring and $\thicksim$ is a congruence
relation over $S$. Then $S/\thicksim$ is also a (c-)semiring and
the natural homomorphism $\nu: S\ra S/\thicksim$ is a semiring
homomorphism.
\end{ex}

\begin{ex}[projection]\label{ex:proj}
Let $S=\prod_{j\in J}S_j$ be the Cartesian product of a set of
(c-)semirings. Clearly, $S$ itself is also a (c-)semiring. For each
$j\in J$, the $j$-th projection $p_j: S\ra S_j$ is a semiring
homomorphism.
\end{ex}

\subsection{Soft constraints}\label{sec:soft}
\begin{defi}[constraint system \cite{BMR97}] A
\emph{constraint system} is a tuple $CS=\langle S,D,V\rangle$,
where $S$ is a c-semiring, $D$ is a finite set, and $V$ is an
(possibly infinite) ordered set of variables.
\end{defi}
\begin{defi}[type]
Given a constraint system $CS=\langle S,D,V\rangle$. A \emph{type}
is a finite ordered subset of $V$. We write
$\mathfrak{T}=\{\tau\subseteq V: \tau\ \mbox{is\ finite}\}$ for
the set of types.
\end{defi}

\begin{defi}[constraints \cite{BMR97}] Given a
constraint system $CS=\langle S,D,V\rangle$, where $S=\langle
S,+,\times,{\bf 0},{\bf 1}\rangle$, a constraint over $CS$ is a
pair $\langle\dfn, \con\rangle$ where
\begin{itemize}
\item $\con$ is a finite subset of $V$, called the \emph{type} of
the constraint;

\item $ \dfn: D^k\ra S$ is called the \emph{value} of the
constraint, where $k=|\con|$ is the cardinality of $\con$.
\end{itemize}
\end{defi}
In the above definition, if $\dfn: D^k\ra S$ is the maximal
constant function, namely $\dfn(t)={\bf 1}$ for each $k$-tuple
$t$, we call $\langle\dfn,\con\rangle$ the \emph{trivial}
constraint with type $\con$.
\begin{defi}[constraint ordering \cite{BMR97}]\label{def:cx}
For two constraints $c_1=\langle \dfn_1, \con\rangle$ and
$c_2=\langle\dfn_2, \con\rangle$ with type $\con$ over $CS=\langle
S,D,V\rangle$, we say $c_1$ is \emph{constraint below}
{\label{pagecx}} $c_2$, noted as $c_1\cx_S c_2$, if for all
$|\con|$-tuples $t$, $\dfn_1(t)\leq_S \dfn_2(t)$.
\end{defi}
This relation can be extended to sets of constraints in an obvious
way. \label{cx} Given two (possibly infinite) sets of constraints
$C_1$ and $C_2$, assuming that both contain no two constraints of
the same type, we say $C_1$ is \emph{constraint below} $C_2$,
noted as $C_1\cx_S C_2$, if for each type $\con\subseteq V$ one of
the following two conditions holds:
\begin{itemize}
\item[(1)] There exist two constraints $c_1$ and $c_2$ with type
$\con$ in $C_1$ and $C_2$ respectively, such that $c_1\cx_S c_2$;

\item[(2)]$C_2$ contains no constraints of type $\con$, or $C_2$
contains the trivial constraint of type $\con$.
\end{itemize}
Two sets of constraints $C_1$ and $C_2$ are called
(\emph{constraint}) \emph{equal}, if $C_1\cx_S C_2$ and $C_2\cx_S
C_1$. In this case, we write $C_1=C_2$. This definition is in
accordance with the basic requirement that adding to a set of
constraints $C$ a trivial constraint should not change the meaning
of $C$.

\begin{defi}[soft constraint problem \cite{BMR97}]
 Given a constraint system $CS=\langle
S,D,V\rangle$, a soft constraint satisfaction problem (SCSP) over
$CS$ is a pair $\langle C,\con\rangle$, where $C$ is a finite set
of constraints, and $\con$, the type of the problem, is a finite
subset of $V$. We assume that no two constraints with the same
type appear in $C$.
\end{defi}

Naturally, given two SCSPs $P_1=\langle C_1, \con\rangle$ and
$P_2=\langle C_2,\con\rangle$, we say $P_1$ is \emph{constraint
below} $P_2$, noted as $P_1\cx_S P_2$, if $C_1\cx_S C_2$. Also,
$P_1$ and $P_2$ are said to be (\emph{constraint}) \emph{equal}, if
$C_1$ and $C_2$ are constraint equal. In this case, we also write
$P_1=P_2$. We call this the \emph{constraint ordering} on sets of
SCSPs with type $\con$ over $CS$. Clearly, two SCSPs are constraint
equal if and only if they differ only in trivial constraints.

To give a formal description of the solution of an SCSP, we need
two additional concepts.

\begin{defi}[combination \cite{BMR97}]
Given a finite set of constraints
$C=\{\langle\dfn_i,\con_i\rangle: i=1,\cdots,n\}$, their
\emph{combination} $\bigotimes C$ is the constraint
$\langle\dfn,\con\rangle$ defined by
$\con=\bigcup_{i=1}^{n}\con_i$ and $ \dfn(t)=\prod_{i=1}^n
\dfn_i(t|^{\con}_{\con_i})$, where by $t|^{X}_{Y}$ we mean the
projection of tuple $t$, which is defined over the set of
variables $X$, over the set of variables $Y\subseteq X$.
\end{defi}

\begin{defi}[projection \cite{BMR97}]
Given a constraint $c=\langle\dfn,\con\rangle$ and a subset $I$ of
$V$, the \emph{projection} of $c$ over $I$, denoted by
$c\Downarrow_I$, is the constraint
$\langle\dfn^\prime,\con^\prime\rangle$ where
$\con^\prime=\con\cap I$ and $
\dfn'(t^\prime)=\sum\{\dfn(t):t|^{\con}_{\con\cap I}=t^\prime\}$.
Particularly, if $I=\varnothing$, then
$c\Downarrow_\varnothing:\{\varepsilon\}\ra S$ maps 0-tuple
$\varepsilon$ to $\sum\{\dfn(t):t{\rm\ is\ a\ tuple\ with\ type}\
 \con\}$, which is the sum of the values associated to all
$|\con|$-tuples.
\end{defi}
Now the concept of solution can be defined as the projection of
the combination of all constraints over the type of the problem.

\begin{defi}[solution and optimal solution]
The \emph{solution} of an SCSP $P=\langle C,\con\rangle$ is a
constraint of type $\con$ which is defined as:
\begin{equation}
Sol(P)=(c^\ast\times\bigotimes C)\Downarrow_{\con}
\end{equation}
where $c^\ast$ is the maximal constraint with type $\con$.

Write $Sol(P)=\langle\dfn,\con\rangle$, a $|\con|$-tuple $t$ is an
\emph{optimal solution} of $P$ if $ \dfn(t)$ is maximal, that is
to say there is no $t^\prime$ such that
$\dfn(t^\prime)>_S\dfn(t)$. We write $Opt(P)$ for the set of
optimal solutions of $P$. For any $|\con|$-tuple $t$, we also
write $Sol(P)(t)$ for $\dfn(t)$.
\end{defi}
%\begin{rmk}The definition of \emph{solution}
%given here differs from the one given in \cite{BMR97,BCR02} in
%that we require the solution of a problem with type $\con$ to be a
%constraint of type $\con$. Given two constraint problems $P$ and
%$Q$ that are constraint equal, this guarantees that $Sol(P)$ and
%$Sol(Q)$ are the same constraint.
%\end{rmk}

\section{Translation and semiring homomorphism}
Let $S=\langle S,+,\times,{\bf 0}, {\bf 1}\rangle$ and
$\widetilde{S}=\langle
\widetilde{S},\widetilde{+},\widetilde{\times},\widetilde{\bf 0},
\widetilde{\bf 1}\rangle$ be two c-semirings and let $\alpha: S\ra
\widetilde{S}$ be an arbitrary mapping from $S$ to
$\widetilde{S}$. Also let $D$ be a nonempty finite set and let $V$
be an ordered set of variables. Fix a type $\con\subseteq V$. We
now investigate the relation between problems over $S$ and those
over $\widetilde{S}$.

\begin{defi}[translation]\label{def:abs} \hypertarget{abs}
Let $P=\langle C,\con\rangle$ be an SCSP over $S$ where
$C=\{c_0,\cdots,c_n\}$, $c_i=\langle \dfn_i, \con_i\rangle$, and
$\dfn_i: D^{|\con_i|}\ra S$. By applying $\alpha$ to each
constraints respectively, we get an SCSP
$\langle\widetilde{C},\con\rangle$ over $\widetilde{S}$, called
the $\alpha$-\emph{translated problem} of $P$, which is defined by
$\widetilde{C}=\{\widetilde{c_1}\cdots\widetilde{c_n}\}$,
$\widetilde{c_i}=\langle\widetilde{\dfn_i},\con_i\rangle$, and
$\widetilde{\dfn_i}=\alpha\circ \dfn_i:
D^{|\con_i|}\ra\widetilde{S}$.
\begin{equation*}
\begin{CD}
D^{|\con_i|}         @>\dfn_i>> S\\
@V{\widetilde{\dfn_i}}VV      @VV{\alpha}V\\
\widetilde{S}         @= \widetilde{S}
\end{CD}
\end{equation*}
We write $\alpha(P)$ for the $\alpha$-translated problem of $P$.
\end{defi}

Without loss of generality, in what follows we assume $\alpha({\bf
0})=\widetilde{\bf 0}$, and $\alpha({\bf 1})=\widetilde{\bf 1}$.
We say $\alpha$ \emph{preserves problem ordering}, if for any two
SCSPs $P,Q$ over $S$, we have
\begin{equation}\label{eq:pop}
Sol(P)\cx_S Sol(Q)\Rightarrow
Sol(\alpha(P))\cx_{\widetilde{S}}Sol(\alpha(Q))
\end{equation}
The following theorem then characterizes when $\alpha$ preserves
problem ordering.
\begin{thm}
Let $\alpha$ be a mapping from c-semiring $S$ to c-semiring
$\widetilde{S}$ such that $\alpha({\bf 0})=\widetilde{\bf 0}$,
$\alpha({\bf 1})=\widetilde{\bf 1}$. Suppose $D$ contains more
than two elements and $k=|\con|>0$. Then $\alpha$ preserves
problem ordering if and only if $\alpha$ is a semiring
homomorphism, that is, for all $a,b\in S$, $\alpha(a\times
b)=\alpha(a)\widetilde{\times}\alpha(b)$,
$\alpha(a+b)=\alpha(a)\widetilde{+}\alpha(b)$.
\end{thm}
\begin{proof} Note that if $\alpha$ preserves $+$ and $\times$, then
$\alpha$ commutes with operators $\prod$ and $\sum$. Clearly
$\alpha$ is also monotonic. Hence, by definition of solution,
$\alpha$ preserves problem ordering.

On the other hand, suppose $\alpha$ preserves problem ordering. We
first prove $\alpha(a+b)=\alpha(a)\widetilde{+}\alpha(b)$  for
$a,b\in S$. We show this by construction.

\begin{scsp}\label{scsp:+}
Suppose $\con=\{y_1,y_2,\cdots,y_k\}$. Take $c_i=\langle
\dfn_i,\con_i\rangle$ with $\con_i=\{x_1,x_2\}$ (i=1,2), where
$x_2\in \con$, $x_1\not\in \con$ and
\[\begin{array}{lccl}
\dfn_1: D^2\ra S{\hspace*{2em}}   & (x_1,x_2)&\mapsto& {a}\ {\rm if}\ x_1=x_2,\\
                   & (x_1,x_2)&\mapsto& {b}\ {\rm if}\ x_1\not=x_2,\\
\dfn_2: D^2\ra S   & (x_1,x_2)&\mapsto& {a+b}.
\end{array}\]
Set $P=\langle \{c_1\},\con\rangle$ and $Q=\langle
\{c_2\},\con\rangle$. Then for each $k$-tuple $(y_1,\cdots,y_k)$,
$Sol(P)(y_1$, $\cdots,y_k)=a+b=Sol(Q)(y_1,\cdots,y_k)$. By the
assumption that $\alpha$ preserves problem ordering, we have
\[\alpha(a)\widetilde{+}\alpha(b)=Sol(\widetilde{P})(y_1,\cdots,y_k)
=Sol(\widetilde{Q})(y_1,\cdots,y_k)=\alpha(a+b).\]
\end{scsp}
Next, we prove $\alpha(a\times
b)=\alpha(a)\widetilde{\times}\alpha(b)$ for $a,b\in S$. We also
show this by construction.
\begin{scsp}\label{scsp:times}
Suppose $\con=\{y_1,y_2,\cdots,y_k\}$. Take $c_1=\langle
\dfn_1,\{x\}\rangle$, $c_2=\langle \dfn_2,\con\rangle$ and
$c_3=\langle \dfn_3,\con\rangle$,  where $x\not\in \con$ and
\[\begin{array}{lccl}
\dfn_1: D\ra S                    & x&\mapsto& {a},\\
\dfn_2: D^{k}\ra S{\hspace*{1cm}} & (y_1,\cdots,y_k)&\mapsto& {b},\\
\dfn_3: D^{k}\ra S
&(y_1,\cdots,y_k)&\mapsto&{a\times b}
\end{array}\]
Set $P=\langle \{c_1,c_2\},\con\rangle$ and $Q=\langle
\{c_3\},\con\rangle$. Then for each $k$-tuple $(y_1,\cdots,y_k)$,
$Sol(P)(y_1,\cdots,y_k)=a\times b=Sol(Q)(y_1,\cdots,y_k)$. By
assumption, we have \[\alpha(a)\widetilde{\times}\alpha(b)
=Sol(\widetilde{P})(y_1,\cdots,y_k)
=Sol(\widetilde{Q})(y_1,\cdots,y_k)=\alpha(a\times b).\]
\end{scsp}
This ends the proof.
\end{proof}

Thus if $\alpha$ is a semiring homomorphism, it preserves problem
ordering. Note that semiring homomorphism also preserves
constraint ordering, i.e. for any two SCSPs $P,Q$ over $S$, we
have
\begin{equation}\label{eq:cop}
P\cx_S Q\Rightarrow\alpha(P)\cx_{\widetilde{S}}\alpha(Q)
\end{equation}

\section{Mappings preserving optimal solutions}
In this section we discuss when a translation preserves optimal
solutions, i.e. when all optimal solutions of the concrete problem
are also optimal in the abstract problem.
\begin{defi}
Let $\alpha: S\ra\widetilde{S}$ be a mapping between two
c-semirings. We say $\alpha$ preserves optimal solutions if
$Opt(P)\subseteq Opt(\alpha(P))$ holds for any SCSP $P$ over $S$.
\end{defi}
The following order-reflecting property plays a key role.
\begin{defi}
Let $(\mathcal{C}, \sqsubseteq)$ and $(\mathcal{A},\leq)$ be two
posets. A mapping $\alpha: \mathcal{C}\ra\mathcal{A}$ is said to be
\emph{order-reflecting} if
\begin{equation}\label{eq:or}
(\forall a,b\in\mathcal{C})\ \alpha(a)<\alpha(b)\Rightarrow
a\sqsubset b
\end{equation}
\end{defi}

In the remainder of this section we show that
$\alpha$ preserves optimal solutions
if and only if $\alpha$ is an order-reflecting semiring
homomorphism. To this end, we need several lemmas.

Recall that $+$ is idempotent and monotonic
on $\leq_S$ for any c-semiring $S=\langle S,+,\times,{\bf 0}, {\bf
1}\rangle$. The following lemma then identifies a necessary and
sufficient condition for $\alpha$ preserving optimal solutions.

\begin{lem} \label{lem:1}
Let $\alpha$ be a mapping from c-semiring $S$ to c-semiring
$\widetilde{S}$ such that $\alpha({\bf 0})=\widetilde{\bf 0},\
\alpha({\bf 1})=\widetilde{\bf 1}$. Then $\alpha$ preserves
optimal solutions for all constraint systems if and only if the
following condition holds for any two positive integers $m,n$:
\begin{equation}\label{eq1}
\widetilde\sum_{i=1}^{n}\widetilde\prod_{j=1}^{m}\alpha(u_{ij})
<_{\widetilde{S}}
\widetilde\sum_{i=1}^{n}\widetilde\prod_{j=1}^{m}\alpha(v_{ij})\Rightarrow
\sum_{i=1}^{n}\prod_{j=1}^{m}u_{ij} <_S
\sum_{i=1}^{n}\prod_{j=1}^{m}v_{ij}.
\end{equation}
\end{lem}

\begin{proof} Suppose that $\alpha$ satisfies the above
Equation~\ref{eq1}. Given an SCSP $P=\langle C,\con\rangle$ over
$S$ with $C=\{c_i\}_{i=1}^m$ and $c_i=\langle
\dfn_i,\con_i\rangle$. Take a tuple $t$ that is optimal in $P$. We
now show $t$ is also optimal in $\alpha(P)$.

Set $\Con=\con\cup\bigcup_{k=1}^{m} \con_k$. Take
$T(t)=\{t^\prime:\ t^\prime|^{\Con}_{\con}=t\}$. Set $n=|T(t)|$
and write $T(t)=\{t_i: 1\leq i\leq n\}$. For each $1\leq i\leq n$
and each $1\leq j\leq m$, set $u_{ij}=c_j(t_i|^{\Con}_{\con_j})$.
Then
\[u=Sol(P)(t)=\sum_{t_i\in T(t)}\prod_{j=1}^{m}c_j(t_i|^{\Con}_{\con_j})
=\sum_{i=1}^{n}\prod_{j=1}^{m}u_{ij},\] and
\[\widetilde{u}=Sol(\alpha(P))(t)
=\widetilde\sum_{i=1}^{n}\widetilde\prod_{j=1}^{m}\alpha(u_{ij}).\]
Suppose $t$ is not optimal in $\alpha(P)$. Then there exists some
$\bar{t}$ that has value
$\widetilde{v}>_{\widetilde{S}}\widetilde{u}$ in ${\alpha}(P)$.
Notice that $T(\bar{t})=\{t^\prime:\
t^\prime|^{\Con}_{\con}=\bar{t}\}$ also has $n=|T(t)|$ elements.
Similarly we can write
\[v=\sum_{i=1}^{n}\prod_{j=1}^{m}v_{ij}\] for the value of $\bar{t}$
in $P$. Now since
\[\widetilde\sum_{i=1}^{n}\widetilde\prod_{j=1}^{m}\alpha(u_{ij})=
\widetilde{u}<_{\widetilde{S}}\widetilde{v}
=\widetilde\sum_{i=1}^{n}\widetilde\prod_{j=1}^{m}\alpha(v_{ij}),\]
entreating Equation~\ref{eq1}, we have $u<_S v$. This contradicts
the assumption that $t$ is optimal in $P$ with value $u$.

On the other hand, suppose that $\alpha$ preserves optimal
solutions. By contradiction, suppose Equation~\ref{eq1} doesn't
hold. That is, we have some
$u=\sum_{i=1}^{n}\prod_{j=1}^{m}u_{ij}$ and
$v=\sum_{i=1}^{n}\prod_{j=1}^{m}v_{ij}$ such that
\[u\nless_S v,\
 \ \widetilde{u}=\widetilde\sum_{i=1}^{n}\widetilde\prod_{j=1}^{m}\alpha(u_{ij})
 <_{\widetilde{S}}
\widetilde\sum_{i=1}^{n}\widetilde\prod_{j=1}^{m}\alpha(v_{ij})=\widetilde{v}.\]
Our next example shows that this is impossible.

\begin{scsp}
Take $D=\{d_1,d_2,\cdots,d_n\}$, $V=\{x_0,x_1,\cdots,x_{n}\}$, and
$\con=\{x_0\}$. For $1\leq j\leq m$, set $\con_j=V-\{x_j\}$, and
define $\dfn_j: D^n\ra S$ as follows:
\[
\dfn_j (x_0,y_2,\cdots,y_n) =\left\{
\begin{array}{cl}
          u_{ij},   & {\rm if}\ \ x_0=d_1\
                    {\rm and}\ y_2=\cdots=y_n=d_i, \\
          v_{ij},   & {\rm if}\ \ x_0=d_2\
                    {\rm and}\ y_2=\cdots=y_n=d_i, \\
         {\bf 0},   & {\rm otherwise}.
\end{array}
             \right.
\]
Set $C=\{\langle\dfn_j,\con_j\rangle\}_{j=1}^m$. Consider now the
SCSP $P=\langle C,\con\rangle$. Then the two 1-tuples $t=(d_1)$
and $t^\prime=(d_2)$ have values
$u=\sum_{i=1}^{n}\prod_{j=1}^{m}u_{ij}$ and
$v=\sum_{i=1}^{n}\prod_{j=1}^{m}v_{ij}$ respectively in $P$.
Applying $\alpha$ to $P$, we have an SCSP $\alpha(P)$ over
$\widetilde{S}$. Recall $\alpha({\bf 0})=\widetilde{\bf 0}$. In
the new problem, $t$ and $t^\prime$ have values
$\widetilde{u}=\widetilde\sum_{i=1}^{n}\widetilde\prod_{j=1}^{m}\alpha(u_{ij})$
and
$\widetilde{v}=\widetilde\sum_{i=1}^{n}\widetilde\prod_{j=1}^{m}\alpha(v_{ij})$
respectively. Since $t$ is an optimal solution of $P$, by the
assumption that $\alpha$ preserves optimal solutions, $t$ is also
an optimal solution of $\alpha(P)$. Recall that
$Sol(\alpha(P))(t)=\widetilde{u}<_{\widetilde{S}}\widetilde{v}
=Sol(\alpha(P))(t^\prime)$. $t$ cannot be optimal in $\alpha(P)$.
This gives a contradiction.
\end{scsp}
As a result, $\alpha$ preserves optimal solutions only if it
satisfies Equation~\ref{eq1}.
\end{proof}

It is easy to show that if $\alpha$ preserves optimal solutions,
then $\alpha$ is order-reflecting.
\begin{lem}\label{lem:2}
Let $\alpha$ be a mapping from c-semiring $S$ to c-semiring
$\widetilde{S}$ such that $\alpha({\bf 0})=\widetilde{\bf 0},\
\alpha({\bf 1})=\widetilde{\bf 1}$. Suppose $\alpha:S\ra
\widetilde{S}$ preserves optimal solutions. Then $\alpha$ is
order-reflecting, that is, for all $u,v\in S$,
$\alpha(u)<_{\widetilde{S}}\alpha(v)$ holds only if  $u<_S v$.
\end{lem}
\begin{proof}
By Lemma~\ref{lem:1}, we know $\alpha$ satisfies
Equation~\ref{eq1} of Lemma~\ref{lem:1}. Taking $m=n=1$, we know
$\alpha$ is order-reflecting.
\end{proof}
The next lemma shows that $\alpha$ preserves optimal solutions
only if it is a semiring homomorphism.
\begin{lem}\label{lem:3}
Let $\alpha$ be a mapping from c-semiring $S$ to c-semiring
$\widetilde{S}$ such that $\alpha({\bf 0})=\widetilde{\bf 0},\
\alpha({\bf 1})=\widetilde{\bf 1}$. Suppose $\alpha:S\ra
\widetilde{S}$ preserves optimal solutions. Then $\alpha$ is a
semiring homomorphism.
\end{lem}
\begin{proof}
By Lemma~\ref{lem:1}, we know $\alpha$ satisfies
Equation~\ref{eq1}. We first show that $\alpha$ is monotonic. Take
$u,v\in S$, $u\leq_S v$. Suppose
$\alpha(u)\not\leq_{\widetilde{S}} \alpha(v)$. Then
$\alpha(v)\widetilde{+}\alpha(v)=\alpha(v)
<_{\widetilde{S}}\alpha(u)\widetilde{+}\alpha(v)$. By
Equation~\ref{eq1}, we have $v=v+v<_S u+v=v$. This is a
contradiction, hence we have
$\alpha(u)\leq_{\widetilde{S}}\alpha(v)$.

Next, for any $u,v\in S$, we show
$\alpha(u+v)=\alpha(u)\widetilde{+}\alpha(v)$. Since $\alpha$ is
monotonic, we have $\alpha(u+v)\geq_{\widetilde{S}}
\alpha(u)\widetilde{+}\alpha(v)$. Suppose
$\alpha(u+v)\widetilde{+}\alpha(u+v)=\alpha(u+v)
>_{\widetilde{S}}\alpha(u)\widetilde{+}\alpha(v)$.
By Equation~\ref{eq1} again, we have $(u+v)+(u+v)>_S u+v$, also a
contradiction.

Finally, for $u,v\in S$, we show $\alpha(u\times
v)=\alpha(u)\widetilde{\times}\alpha(v)$. Suppose not and set
$w=\alpha(u)\widetilde{\times}\alpha(v)\widetilde{+}\alpha(u\times
v)$. Then we have either
$\alpha(u)\widetilde{\times}\alpha(v)<_{\widetilde{S}}w$ or
$\alpha(u\times v)<_{\widetilde{S}}w$. Since $\alpha({\bf
0})=\widetilde{\bf 0}$, $\alpha({\bf 1})=\widetilde{\bf 1}$, these
two inequalities can be rewritten respectively as
\[\alpha(u)\widetilde{\times}\alpha(v)+\alpha({\bf
1})\widetilde{\times}\alpha({\bf 0})
<_{\widetilde{S}}\alpha(u)\widetilde{\times}\alpha(v)\widetilde{+}\alpha(u\times
v)\widetilde{\times}\alpha(\widetilde{\bf 1})\] \noindent and
\[\alpha({\bf
1})\widetilde{\times}\alpha({\bf 0})+\alpha(u\times
v)\widetilde{\times}\alpha({\bf 1})
<_{\widetilde{S}}\alpha(u)\widetilde{\times}\alpha(v)\widetilde{+}\alpha(u\times
v)\widetilde{\times}\alpha(\widetilde{\bf 1}).\] \noindent By
Equation~\ref{eq1} again, we have either $u\times v+{\bf
1}\times{\bf 0} <_S u\times v+(u\times v)\times{\bf 1}$ or ${\bf
1}\times {\bf 0}+ (u\times v)\times{\bf 1}<_S u\times v+ (u\times
v)\times{\bf 1}$. Both give rise to a contradiction. This ends the
proof.
\end{proof}

We now achieve our main result:

\begin{thm} \label{mythm27}
Let $\alpha$ be a mapping from c-semiring $S$ to c-semiring
$\widetilde{S}$ such that $\alpha({\bf 0})=\widetilde{\bf 0},\
\alpha({\bf 1})=\widetilde{\bf 1}$. Then $\alpha$ preserves
optimal solutions for all constraint systems if and only if
$\alpha$ is an order-reflecting semiring homomorphism.
\end{thm}

\begin{proof} The necessity part of the
theorem follows from Lemmas~\ref{lem:2} and \ref{lem:3}. As for
the sufficiency part, we need only to show that, if $\alpha$ is an
order-reflecting semiring homomorphism, then $\alpha$ satisfies
Equation~\ref{eq1}. Suppose
\[\widetilde\sum_{i=1}^{n}\widetilde\prod_{j=1}^{m}\alpha(u_{ij})
<_{\widetilde{S}}\widetilde\sum_{i=1}^{n}\widetilde\prod_{j=1}^{m}\alpha(v_{ij}).\]
Clearly we have
\[\alpha(\sum_{i=1}^{n}\prod_{j=1}^{m}u_{ij})=
\widetilde\sum_{i=1}^{n}\widetilde\prod_{j=1}^{m}\alpha(u_{ij})
<_{\widetilde{S}}
\widetilde\sum_{i=1}^{n}\widetilde\prod_{j=1}^{m}\alpha(v_{ij})=
\alpha(\sum_{i=1}^{n}\prod_{j=1}^{m}v_{ij})\] \noindent since
$\alpha$ commutes with $\sum$ and $\prod$. By order-reflecting, we
have immediately
\[\sum_{i=1}^{n}\prod_{j=1}^{m}u_{ij}<_S\sum_{i=1}^{n}\prod_{j=1}^{m}v_{ij}.\]
This ends the proof.
\end{proof}

\section{Computing concrete optimal
solutions from abstract ones}

In the above section, we investigated conditions under which
\emph{all} optimal solutions of concrete problem can be related
\emph{precisely} to those of abstract problem. There are often
situations where it suffices to find \emph{some} optimal
solutions or simply a good approximation of the concrete optimal
solutions. This section shows that, even without the
order-reflecting condition, semiring homomorphism can be used to
find some optimal solutions of concrete problem using abstract
ones.

\begin{thm}\label{mythm31}
Let $\alpha:S\ra\widetilde{S}$ be a semiring homomorphism. Given
an SCSP $P$ over $S$, suppose $t\in Opt(\alpha(P))$ has value $v$
in $P$ and value $\widetilde{v}$ in $\alpha(P)$. Then there exists
$\bar{t}\in Opt(P)\cap Opt(\alpha(P))$ with value $\bar{v}\geq_S
v$ in $P$ and value $\widetilde{v}$ in $\alpha(P)$. Moreover, we
have $\alpha(\bar{v})=\alpha(v)=\widetilde{v}$.
\end{thm}

\begin{proof} Suppose $P=\langle C,\con\rangle$,
$C=\{c_i\}_{i=1}^m$ and $c_i=\langle \dfn_i,\con_i\rangle$. Set
$\Con=\con\cup\bigcup\{\con_j\}_{j=1}^m$ and $k=|\Con|$. Suppose
$t$ is an optimal solution of $\alpha(P)$, with semiring value
$\widetilde{v}$ in $\alpha(P)$ and $v$ in $P$. By definition of
solution, we have
\[v=Sol(P)(t)=\sum_{t^\prime|_{\con}^{\Con}=t}\prod_{j=1}^m
\dfn_j(t^\prime|_{\con_j}).\] Denote
\[T(t)=\{t^\prime: t^\prime\ \mbox{is\ a\ }|k|\mbox{-tuple\ with}\
t^\prime|_{\con}^{\Con}=t\}.\] Set $n=|T(t)|$, and write
$T=\{t_1,\cdots,t_n\}$. For each $1\leq i\leq n$ and each $1\leq
j\leq m$, set $v_{ij}= \dfn_j(t_i|_{\con_j})$. Then
\[v=\sum_{i=1}^n\prod_{j=1}^m v_{ij},\ \
\widetilde{v}=\widetilde\sum_{i=1}^n\widetilde\prod_{j=1}^m\alpha(v_{ij}).\]

Since $\alpha$ preserves sums and products, we have
\[\alpha(v)
=\alpha(\sum_{i=1}^n\prod_{j=1}^m v_{ij})
=\widetilde\sum_{i=1}^n\alpha(\prod_{j=1}^m v_{ij})
=\widetilde\sum_{i=1}^n\widetilde\prod_{j=1}^m \alpha(v_{ij})
=\widetilde{v}.\] Notice
that if $t$ is also optimal in $P$, then we can choose
$\bar{t}=t$. Suppose $t$ is not optimal in $P$. Then there is a
tuple $\bar{t}$ that is optimal in $P$, say with value
$\overline{v}>_S v$. Denote
\[T(\bar{t})=\{t^\prime: t^\prime\ \mbox{is\ a\
}|k|\mbox{-tuple\ with}\ t^\prime|_{\con}^{\Con}=\bar{t}\}.\]
Clearly $|T(\bar{t})|=|T(t)|=n$. Write
$T(\bar{t})=\{\bar{t}_1,\cdots,\bar{t}_n\}$. For each $1\leq i\leq
n$ and each $1\leq j\leq m$, set $u_{ij}=
\dfn_j(\bar{t}_i|_{\con_j})$. Then
\[\overline{v}=\sum_{i=1}^n\prod_{j=1}^m u_{ij}.\]
Now we show
$\alpha(\overline{v})\leq_{\widetilde{S}}\widetilde{v}$.

By $v<_S\overline{v}$, we have
$\alpha(v)\leq_{\widetilde{S}}\alpha(\overline{v})$. Then
\begin{eqnarray*}
 \widetilde{v}&=&\widetilde\sum_{i=1}^n\widetilde\prod_{j=1}^m
\alpha(v_{ij})\\
&=&\alpha(\sum_{i=1}^n\prod_{j=1}^m
v_{ij}) \\
&=&\alpha(v)\leq_{\widetilde{S}}\alpha(\overline{v})
=\alpha(\sum_{i=1}^n\prod_{j=1}^m
u_{ij})=\widetilde\sum_{i=1}^n\widetilde\prod_{j=1}^m
\alpha(u_{ij})=\widetilde{\overline{v}}
\end{eqnarray*}
where the last term, $\widetilde{\overline{v}}$, is the value of
$\bar{t}$ in $\alpha(P)$. Now since $t$ is optimal in $\alpha(P)$,
we have $\widetilde{v}=\alpha(v)=
\alpha(\overline{v})=\widetilde{\overline{v}}$. That is, $\bar{t}$
is also optimal in $\alpha(P)$ with value $\widetilde{v}$.
\end{proof}
\begin{rmk}
If our aim is to find some instead of all optimal solutions of the
concrete problem $P$, by Theorem~\ref{mythm31} we could first find
all optimal solutions of the abstract problem $\alpha(P)$, and
then compute their values in $P$, tuples that have maximal values
in $P$ are optimal solutions of $P$. In this sense, this theorem
is more desirable than Theorem~\ref{mythm27} because we do not
need the assumption that $\alpha$ is order-reflecting.

Theorem~\ref{mythm31} can also be applied to find good approximations
of the optimal solutions of $P$. Given an optimal solution $t\in
Opt(\alpha(P))$ with value $\tilde{v}\in\widetilde{S}$, then by
Theorem~\ref{mythm31} there is an optimal solution $\bar{t}\in
Opt(P)$ with value in the set $\{u\in S:\alpha(u)=\widetilde{v}\}$.
\end{rmk}

Note that Theorem~\ref{mythm31} requires $\alpha$ to be a semiring
homomorphism. This condition is still a little restrictive. Take the
probabilistic semiring
$S_{prop}=\langle[0,1],\max,\times,0,1\rangle$ and the classical
semiring $S_{CSP}=\langle\{T,F\},\vee,\wedge,F,T\rangle$ as example,
there are no nontrivial homomorphisms between $S_{prop}$ and
$S_{CSP}$. This is because $\alpha(a\times
b)=\alpha(a)\wedge\alpha(b)$ requires $\alpha(a^n)=\alpha(a)$ for
any $a\in[0,1]$ and any positive integer $n$, which implies
$(\forall a>0)\alpha(a)=1$ or $ (\forall a<1)\alpha(a)=1$.

In the remainder of this section, we relax this condition.
\begin{defi}[quasi-homomorphism]
 A mapping $\psi$ from semiring $\langle S, +, \times$,
{\bf 0}, {\bf 1}$\rangle$ to semiring $\langle \widetilde {S},
\widetilde {+}, \widetilde{\times}, \widetilde{\bf 0},
\widetilde{\bf 1}\rangle$ is said to be a \emph{quasi-homomorphism}
if for any $a,b\in S$
\begin{itemize}
\item $\psi(\bf 0)=\widetilde{\bf 0}$, $\psi(\bf 1)=\widetilde{\bf
1}$; and

\item $\psi(a+b)=\psi(a)\widetilde {+}\psi(b)$; and

\item $\psi(a\times b)\leq_{\widetilde{S}}\psi(a)\widetilde {\times}\psi(b)$.
\end{itemize}
\end{defi}
The last condition is exactly the \emph{locally correctness} of
$\widetilde\times$ w.r.t. $\times$ \cite{BCR02}. Clearly, each
monotonic surjective mapping between $S_{prop}$ and $S_{CSP}$ is a
quasi-homomorphism.

The following theorem shows that a quasi-homomorphism is also
useful.

\begin{thm}\label{mythm31+}
Let $\alpha:S\ra\widetilde{S}$ be a quasi-semiring homomorphism.
Given an SCSP $P$ over $S$, suppose $t\in Opt(\alpha(P))$ has value
$v$ in $P$ and value $\widetilde{v}$ in $\alpha(P)$. Then there
exists an optimal solution $\bar{t}$ of $P$, say with value
$\bar{v}\geq_S v$ in $P$, such that
$\alpha(\bar{v})\not>_{\widetilde{S}}\widetilde{v}$.
\end{thm}
\begin{proof}
The proof is straightforward.
\end{proof}
Note that if $\widetilde{S}$ is totally ordered, then the above
conclusion can be rephrased as
$\alpha(\bar{v})\leq_{\widetilde{S}}\widetilde{v}$. But the
following example shows this is not always true.

\begin{figure}[thb]\centering
\fbox{\includegraphics[width=.6\textwidth]{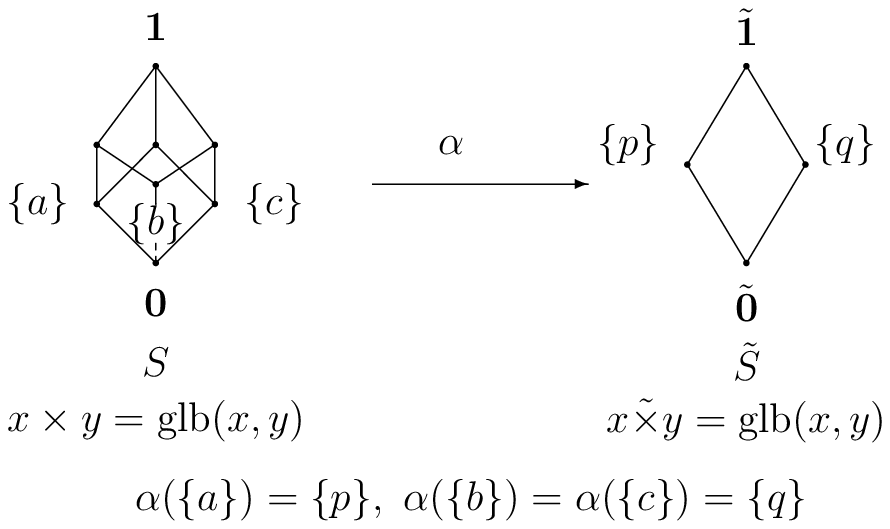}} \caption{A
counter-example} \label{fig:alpha1}
\end{figure}

\begin{scsp}\label{scsp:thm29}
Take $D=\{d_1,d_2\}$, $X=\{a,b,c\}$, $Y=\{p,q\}$ and
$V=\{x_1,x_2\}$. Then $S=\langle 2^X,\cup,\cap,\varnothing,X\rangle$
and $\widetilde{S}=\langle 2^Y,\cup,\cap,\varnothing,Y\rangle$ are
two c-semirings, see Figure \ref{fig:alpha1}. Let
$\alpha:S\ra\widetilde{S}$ be the mapping specified by
$\alpha(\varnothing)=\varnothing$, $\alpha(\{a\})=\{p\}$,
$\alpha(\{b\})=\alpha(\{c\})=\alpha(\{b,c\})=\{q\}$, and
$\alpha(\{a,b\})=\alpha(\{a,c\})=\alpha(X)=Y$. Note that $\alpha$
preserves {\lub}s. Moreover, since $\alpha$ is monotonic, we have
$\alpha(U\cap W)\subseteq\alpha(U)\cap\alpha(W)$ for any
$U,W\subseteq X$. Therefore $\alpha$ is a quasi-homomorphism.

Define $\dfn_i:D\ra S$ {\rm($i=1,2$)} as follows:
\[\dfn_1(d_1)=\{a\},\ \dfn_1(d_2)=\{b\};\]
\[\dfn_2(d_1)=\{a\},\ \dfn_2(d_2)=\{c\};\]

Consider the SCSP $P=\langle C,V\rangle$ with $C=\{c_1,c_2\}$ and
$c_i=\langle \dfn_i,\{x_i\}\rangle$ for $i=1,2$. Then
\[\begin{array}{l}
Sol(P)(d_1,d_1)=\{a\}\cap\{a\}=\{a\},\\
Sol(P)(d_1,d_2)=\{a\}\cap\{c\}=\varnothing\\
Sol(P)(d_2,d_1)=\{b\}\cap\{a\}=\varnothing,\\
Sol(P)(d_2,d_2)=\{b\}\cap\{c\}=\varnothing
\end{array}\]
 and
\[\begin{array}{l}
Sol(\alpha(P))(d_1,d_1)=\{p\}\cap\{p\}=\{p\},\\
Sol(\alpha(P))(d_1,d_2)=\{p\}\cap\{q\}=\varnothing,\\
Sol(\alpha(P))(d_2,d_1)=\{q\}\cap\{p\}=\varnothing,\\
Sol(\alpha(P))(d_2,d_2)=\{q\}\cap\{q\}=\{q\}.
\end{array}\]
Set $t=(d_2,d_2)$. Clearly, $t$ is an optimal solution of
$\alpha(P)$ with value $\{q\}$ in $\alpha(P)$, and value
$\varnothing$ in $P$. Notice that $\bar{t}=(d_1,d_1)$ is the unique
optimal solution of $P$. Since
$\alpha(\{a\})=\{p\}\not\subseteq\{q\}$, there is no optimal
solution $\hat{t}$ of $P$ such that $\alpha(\hat{t})\subseteq\{q\}$.
\end{scsp}

\section{Related work}
Our abstraction framework is closely related to the work of
Bistarelli et al. \cite{BCR02} and de Givry et al. \cite{degivry}.

\subsection{Galois insertion-based abstraction}
Bistarelli et al. \cite{BCR02} proposed a Galois insertion-based
abstraction scheme for soft constraints. The questions investigated
here were studied in \cite{BCR02}. In particular, Theorems~27, 29,
31 of \cite{BCR02} correspond to our Theorems~\ref{mythm27},
\ref{mythm31+}, and \ref{mythm31}, respectively.

We recall some basic notions concerning abstractions used in
\cite{BCR02}.

\begin{defi}[Galois insertion \cite{G}]\label{galois}
Let $(\mathcal{C}, \sqsubseteq)$ and $(\mathcal{A},\leq)$ be two
posets (the concrete and the abstract domain). A \emph{Galois
connection} $\langle\alpha,\gamma\rangle:(\mathcal{C},
\sqsubseteq)\rightleftarrows(\mathcal{A},\leq)$ is a pair of
monotonic mappings $\alpha:\mathcal{C}\ra\mathcal{A}$ and $\gamma:
\mathcal{A}\ra\mathcal{C}$ such that
\begin{equation}
(\forall x\in\mathcal{C})(\forall y\in\mathcal{A})\ \alpha(x)\leq
y\Leftrightarrow x\sqsubseteq\gamma(y)
\end{equation}
In this case, we call $\gamma$ the upper adjoint (of $\alpha$),
and $\alpha$ the lower adjoint (of $\gamma$).  A Galois connection
$\langle\alpha,\gamma\rangle: (\mathcal{C},
\sqsubseteq)\rightleftarrows(\mathcal{A},\leq)$ is called a
\emph{Galois insertion} (of $\mathcal{A}$ in $\mathcal{C}$) if
$\alpha\circ\gamma=id_\mathcal{A}$.
\end{defi}

\begin{defi}[abstraction]\label{abstraction}
A mapping $\alpha:S\ra\widetilde{S}$ between two c-semirings is
called an \emph{abstraction} if
\begin{itemize}
\item [1.] $\alpha$ has an upper adjoint $\gamma$ such that
$\langle \alpha,\gamma\rangle:S\rightleftharpoons\widetilde{S}$ is
a Galois insertion

\item [2.] $\widetilde\times$ is \emph{locally correct} with
respect to $\times$, i.e. $(\forall a,b\in S)\ \alpha(a\times
b)\leq_{\widetilde{S}}\alpha(a)\widetilde{\times}\alpha(b)$.
\end{itemize}
\end{defi}

Theorem~27 of \cite{BCR02} gives a sufficient condition for a
Galois insertion preserving optimal solutions. This condition,
called \emph{order-preserving}, is defined as follows:

\begin{defi}[\cite{BCR02}]\label{def:op}
Given a Galois insertion $\langle\alpha,\gamma\rangle:
S\rightleftarrows\widetilde{S}$, $\alpha$ is said to be
\emph{order-preserving} if for any two sets $I_1$ and $I_2$, we
have
\begin{equation}\label{eq2}
\widetilde\prod_{x\in
I_1}\alpha(x)\leq_{\widetilde{S}}\widetilde\prod_{x\in
I_2}\alpha(x)\Rightarrow\prod_{x\in I_1}x\leq_S\prod_{x\in I_2}x.
\end{equation}
\end{defi}

This notion plays an important role in \cite{BCR02}. In fact,
several results (\cite[Theorems~27, 39, 40, 42]{BCR02}) require
this property. The next proposition, however, shows that this
property is too restrictive, since an order-preserving Galois
insertion is indeed a semiring isomorphism.

\begin{pro}\label{prop:op}
Suppose $\langle\alpha,\gamma\rangle:
S\rightleftarrows\widetilde{S}$ is a Galois insertion. Then
$\alpha$ is order-preserving if and only if it is a semiring
isomorphism.
\end{pro}
\begin{proof}
The sufficiency part is clear, and we now show the necessity part.
Notice that $\alpha$, as a Galois connection, is monotonic. On the
other hand, given $x,y\in S$, suppose
$\alpha(x)\leq_{\widetilde{S}}\alpha(y)$. By Equation~\ref{eq2},
we have $x\leq_S y$. That is to say, for any $x,y\in S$,
$\alpha(x)\leq_{\widetilde{S}}\alpha(y)$ if and only if $x\leq_S
y$. In particular, $\alpha(x)=\alpha(y)$ implies $x=y$.
This means that $\alpha$ is injective. Moreover,
by definition of Galois insertion, $\alpha$ is also surjective.
Therefore $\alpha$ is an order isomorphism. As a consequence, it
preserves sums.

We next show $\alpha$ preserves products. For $x,y\in S$,
since $\alpha$ is surjective, we
have some $z\in S$ with
$\alpha(z)=\alpha(x)\widetilde{\times}\alpha(y)$. Applying the
order-preserving property, we have $z=x\times y$, hence
$\alpha(x\times
y)=\alpha(z)=\alpha(x)\widetilde{\times}\alpha(y)$, i.e. $\alpha$
preserves products. In summary, $\alpha$ is a semiring
isomorphism.
\end{proof}
Theorem~29 of \cite{BCR02} concerns that, given an optimal solution
of the abstract problem, how to find a reasonable estimation for an
optimal solution of the concrete problem. Let
$\alpha:S\ra\widetilde{S}$ be an abstraction. Given an SCSP $P$ over
$S$, suppose $t$ is an optimal solution of $\alpha(P)$, with
semiring value $\widetilde{v}$ in $\alpha(P)$ and $v$ in $P$. Then
\cite[Theorem~29]{BCR02} asserts that there exists an optimal
solution $\bar{t}$ of $P$, say with value $\overline{v}$, such that
$v\leq\overline{v}\leq\gamma(\widetilde{v})$.

Our Soft Problem \ref{scsp:thm29}, however, shows that
\cite[Theorem~29]{BCR02} is only conditionally true. This is because
the quasi-homomorphism $\alpha$ given there is also an abstraction.
Since each abstraction is also a quasi-homomorphism,
Theorem~\ref{mythm31+} holds for any abstraction.

Our Theorem~\ref{mythm31} corresponds to Theorem~31 of \cite{BCR02},
where the authors consider abstractions between totally ordered
semirings with idempotent multiplicative operations. By
Example~\ref{ex:tom}, we know such an abstraction must be a
homomorphism. Therefore our result is more general than
\cite[Theorem~31]{BCR02}.

\subsection{Aggregation compatible mapping}

There is another abstraction scheme \cite{degivry} for soft
constraints that is closely related to ours, where \emph{valued
CSPs} \cite{Schiex95} are abstracted in order to produce good lower
bounds for the optimal solutions.
\begin{defi}[\cite{degivry}]
A translation $\alpha:S\ra\widetilde{S}$ between two totally ordered
semirings is said to be \emph{aggregation compatible} if
\begin{itemize}
\item [(1)] $\alpha$ is monotonic and $\alpha({\bf
0})=\widetilde{\bf 0},\ \alpha({\bf 1})=\widetilde{\bf 1}$; and

\item [(2)] For any two sets $I_1$ and $I_2$, we
have\footnote{Note that in Equation~\ref{eq4} we replace the two
$\geq$ in Definition~2 of \cite{degivry} with $\leq$. This is
because we should reverse the order of the valuation set $S$ such
that the aggregation operator $\circledast$ is a product operator.}
\begin{equation}\label{eq4}
\alpha(\prod_{x\in I_1}x)\leq_{\widetilde{S}}\alpha(\prod_{x\in
I_2}x)\Rightarrow \widetilde\prod_{x\in
I_1}\alpha(x)\leq_{\widetilde{S}}\widetilde\prod_{x\in
I_2}\alpha(x).
\end{equation}
\end{itemize}
\end{defi}
The next theorem shows that an aggregate compatible mapping must be
a semiring homomorphism.

\begin{thm} Let $\alpha:S\ra\widetilde{S}$ be a mapping between two totally ordered
semirings. Then $\alpha$ is aggregate compatible if and only if
$\alpha$ is a semiring homomorphism.
\end{thm}
\begin{proof}
 A semiring homomorphism is clearly aggregate compatible.
On the other hand, suppose $\alpha$ is aggregate compatible. Since
it is monotonic, $\alpha$ preserves sums. Moreover, by
Equation~\ref{eq4}, for any $a,b\in S$, taking $I_1=\{a,b\}$,
$I_2=\{a\times b\}$, from $\alpha(a\times b)=\alpha(a\times b)$ we
have $\alpha(a)\widetilde{\times}\alpha(b)=\alpha(a\times b)$. That
is, $\alpha$ also preserves products. Hence $\alpha$ is a semiring
homomorphism.
\end{proof}
Therefore our framework is also a generalization of that of de Givry
et al. More importantly, results obtained in Sections 4 and 5 can be
applied to valued CSPs.

We first note that any monotonic mapping from a totally ordered set
is order-reflecting.
\begin{lem}\label{lemma:tom}
Let $(\mathcal{C}, \sqsubseteq)$ be a totally ordered set, and
$(\mathcal{A},\leq)$ a poset. Suppose
$\alpha:\mathcal{C}\ra\mathcal{A}$ is monotonic mapping. Then
$\alpha$ is order-reflecting.
\end{lem}
\begin{proof}
By contradiction, suppose there are $a,b\in\mathcal{C}$ such that
$\alpha(a)<\alpha(b)$ but $a\not\sqsubset b$. Then since
$\sqsubseteq$ is a total order we know $b\sqsubseteq a$. But by
the monotonicity of $\alpha$, we have $\alpha(b)\leq\alpha(a)$.
This contradicts the assumption that $\alpha(a)<\alpha(b)$.
Therefore $\alpha$ is order-reflecting.
\end{proof}

Now, we have the following corollary of Theorem~\ref{mythm27}, which
was also obtained by de Givry et al. \cite{degivry} for aggregation
compatible mappings.
\begin{cor}\label{cor1}
Let $\alpha$ be a semiring homomorphism between two c-semirings
$S$ and $\widetilde{S}$. Suppose $S$ is a totally ordered
c-semiring. Then for any SCSP $P$ over $S$, it holds that
$Opt(P)\subseteq Opt(\alpha(P))$.
\end{cor}
\begin{proof}
By Lemma~\ref{lemma:tom}, $\alpha$ is order-reflecting. The
conclusion then follows directly from Theorem~\ref{mythm27}.
\end{proof}

\section{Conclusions}

In this paper we proposed a homomorphism based abstraction scheme
for soft constraints. The intuition is that we first work in the
abstract problem, finding all optimal solutions, and then use them
to find optimal solutions of the concrete problem. Surprisingly, our
framework turns out to be a generalization of that of de Givry et
al. \cite{degivry}, where they consider totally ordered sets.

In detail, our Theorem~\ref{mythm27} showed that a mapping preserves
optimal solutions if and only if it is an order-reflecting semiring
homomorphism; and Theorem~\ref{mythm31} showed that, for a semiring
homomorphism $\alpha$ and a problem $P$ over $S$, if $t$ is an
optimal solution of $\alpha(P)$, then there is an optimal solution
of $P$, say $\bar{t}$, such that $\bar{t}$ is also optimal in
$\alpha(P)$ and has the same value as $t$. These results greatly
improved or generalized those obtained in Bistarelli et al.
\cite{BCR02}.


\begin{thebibliography}{200}

\bibitem{BCR02} S. Bistarelli, P. Codognet, F. Rossi,
Abstracting soft constraints: Framework, properties, examples,
{\it Artificial Intelligence} {\bf 139} (2002) 175-211.

\bibitem{B96} S. Bistarelli, H. Fargier, U. Montanari, F. Rossi, T. Schiex,
G. Verfaillie,  Semiring-Based CSPs and Valued CSPs: Basic
Properties and Comparison, in: \emph{Over-Constrained Systems,
Lecture Notes in Computer Science,} Vol.1106, Springer, Berlin,
1996, pp.111-150.

\bibitem{BMR97} S. Bistarelli, U. Montanari, F.
Rossi, Semiring-based constraints solving and optimization,
\emph{Journal of the ACM} {\bf 44} (2) (1997) 201-236.



%\bibitem{Cou77}P. Cousot, R. Cousot, Abstract interpretation:
%A unified lattice model for static analysis of programs by
%construction or approximation of fixpoints, in: \emph{Fourth ACM
%Symposium Principles of Programming Languages}, 1977, pp. 238-252.

\bibitem{degivry}S. de Givry, G. Verfaillie, T. Schiex,  Bounding the
Optimum of Constraint Optimization Problems, in: G. Smolka (Ed.),
 \emph{Proc. CP-97, Lecture Notes in
Computer Science}, Vol.1330,  Springer, Berlin, 1997, pp.405-419.


\bibitem{Dub93} D. Dubois, H. Fargier, H. Prade,
The calculus of fuzzy restrictions as a basis for flexible
constraint satisfaction, in: \emph{Proc. IEEE International
Conference on Fuzzy Systems, IEEE}, 1993, pp. 1131-1136.

\bibitem{Far93} H. Fargier, J. Lang,
Uncertainty in constraint satisfaction problems: A probabilistic
approach, in: \emph{Proc. European Conference on Symbolic and
Qualitative Approaches to Reasoning and Uncertainty (ECSQARU),
Lecture Notes in Computer Science}, Vol. 747, Springer, Berlin,
1993, pp. 97-104.

\bibitem{Fre92} E.C. Freuder, R.J. Wallace,
Partial constraint satisfaction, \emph{Artificial Intelligence}
{\bf 58}(1992) 21-70.

\bibitem{G} G. Gierz, K.H. Hofmann, K. Keimel,
J.D. Lowson, M.W. Mislove, and D.S. Scott, \emph{A Compendium of
Continuous Lattices}, Springer, Berlin, 1980.

\bibitem{Mac92}A.K. Mackworth, Constraint satisfaction,
in: S.C. Shapiro (Ed.), \emph{Encyclopedia of AI}, Vol. 1, 2nd
edition, Wiley, New York, 1992, pp. 285-293.

\bibitem{Mon74} U. Montanari, Networks of constraints: Fundamental
properties and application to picture processing,
\emph{Information Science}, {\bf 7} (1974)95-132.

\bibitem{Ros76} A. Rosenfeld, R. Hummel, and S. Zucker,
Scene labelling by relaxation operations. \emph{IEEE Transactions
on Systems, Man and Cybernetics} {\bf 6}(1976)(6).

\bibitem{Rut94} Zs. Ruttkay,
Fuzzy constraint satisfaction, in: \emph{Proc. 3rd IEEE
International Conference on Fuzzy Systems}, 1994, pp. 1263-1268.

\bibitem{Schiex92} T. Schiex, Possibilistic Constraint Satisfaction
Problems, or ``How to Handle Soft Constraints?", in: \emph{Proc.
UAI-92}, 1992, pp.269-275.

\bibitem{Schiex95}T. Schiex, H. Fargier, G. Verfaillie, Valued
Constraint Satisfaction Problems: Hard and Easy Problems, in:
\emph{Proc. IJCAI-95}, Montreal, Quebec, Morgan Kaufmann, San
Mateo, CA, 1995, pp.631-639.

\bibitem{Shapiro81}
L. Shapiro and R. Haralick,  Structural descriptions and inexact
matching. \emph{IEEE Transactions on Pattern Analysis and Machine
Intelligence} {\bf3}(1981) 504-519.





\end{thebibliography}
\end{document}